\documentclass[twoside,11pt]{article}

\usepackage{jmlr2e}

\usepackage{array}
\usepackage{multirow}
\usepackage{rotating}
\usepackage{graphicx}
\usepackage{hyperref}
\usepackage{amsmath}
\usepackage{paralist}

\firstpageno{1}

\begin{document}

\title{Transferring Knowledge from Text to Predict Disease Onset}


\author{\name Yun Liu, PhD \email liuyun@csail.mit.edu \\
       \name Collin M. Stultz, MD, PhD \email cmstultz@mit.edu \\
       \name John V. Guttag, PhD \email guttag@csail.mit.edu \\
       Massachusetts Institute of Technology, Cambridge, MA, U.S.A.
       \AND
       \name Kun-Ta Chuang, PhD \email ktchuang@mail.ncku.edu.tw \\
       \name Fu-Wen Liang, PhD \email fliang81@gmail.com \\
       \name Huey-Jen Su, PhD \email hjsu@mail.ncku.edu.tw \\
       National Cheng Kung University, Tainan, Taiwan
       } 

\maketitle

\begin{abstract}

In many domains such as medicine, training data is in short supply. In such
cases, external knowledge is often helpful in building predictive models. We
propose a novel method to incorporate publicly available domain expertise to
build accurate models. Specifically, we use word2vec models trained on a
domain-specific corpus to estimate the relevance of each feature's text
description to the prediction problem. We use these relevance estimates to
rescale the features, causing more important features to experience weaker
regularization.

We apply our method to predict the onset of five chronic diseases in the next
five years in two genders and two age groups. Our rescaling approach improves
the accuracy of the model, particularly when there are few positive examples.
Furthermore, our method selects $60\%$ fewer features, easing interpretation
by physicians. Our method is applicable to other domains where feature and
outcome descriptions are available.

\end{abstract}

\vspace{-1mm}
\section{Introduction}
\vspace{-1mm}

In many domains such as medicine, training data is in short supply. The need
to use inclusion or exclusion criteria to select the population of interest,
and the scarcity of many outcomes of interest shrink the available data and
compound the problem. A common technique in these scenarios is to leverage
transfer learning from \emph{source data} for related prediction tasks or
populations (\cite{lee2012adapting,wiens2014study,gong2015instance}). However,
obtaining enough useful source data is often difficult.

\begin{figure}[!tb]
\centering
\includegraphics[width=4.0in]{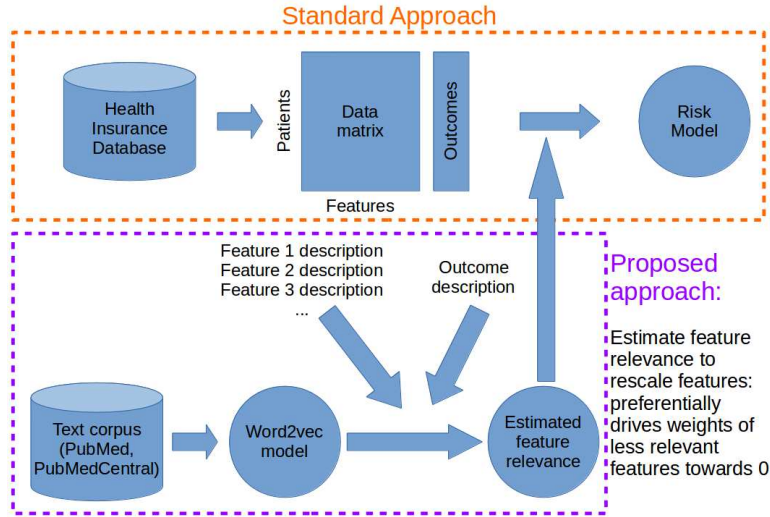}
\vspace{-5mm}

\caption{Overview of transferring knowledge from text to predict outcomes
using a structured dataset.}

\label{w2v_overview}
\end{figure}

However, the target task often contains meta-data, such as \emph{descriptions}
for each feature, and the outcome of interest. We hypothesized that we can use
domain knowledge and these descriptions to estimate the relevance of each
feature to predicting the outcome. Specifically, we transfered knowledge from
publicly available biomedical text corpuses to estimate these relevances, and
used these estimates to automatically guide feature selection and to improve
predictive performance (Figure~\ref{w2v_overview}). We used this method to
predict the onset of various diseases using medical billing records. Our
method substantially improved the Area Under Curve (AUC), and selected fewer
features.

The novelty of our work lies in the use of feature and outcome
\emph{descriptions} and models learned from publicly available auxiliary text
corpuses to improve predictive modeling when data are scarce. This framework
can be further improved for better relevance estimates and can also be applied
in other domains.

The remainder of the paper is arranged as follows. We first describe related
works in transfer learning and supervised learning. Next, we describe the
dataset and outcomes that we examine. Then, we describe our method and our
experimental set up. Finally, we show results and discuss the potential
relevance of our work to other domains.

\vspace{-1mm}
\section{Related Work}
\vspace{-1mm}

Our work is related to transfer learning, which leverages data from a related
prediction task, the source task, to improve prediction on the target task
(\cite{pan2010survey}). Transfer learning has been productively applied to
medical applications such as adapting surgical models to individual hospitals
(\cite{lee2012adapting}), enhancing hospital-specific predictions of
infections (\cite{wiens2014study}), and improving surgical models using data
from other surgeries (\cite{gong2015instance}).

Another body of related work can be found in the field of text classification.
Some authors used expert-coded ontologies such as the Unified Medical Language
System (UMLS) to engineer and extract features from clinical text and used
these features to identify patients with cardiovascular diseases
(\cite{garla2012ontology, yu2015toward}). In non-clinical settings, others
have used ontology in the Open Directory Project
(\cite{gabrilovich2005feature}) and Wikipedia
(\cite{gabrilovich2006overcoming}).

Our work diverges from prior work because our ``target task'' uses structured
medical data, but we transfer knowledge from unstructured external sources to
understand the relationship between the \emph{descriptions} of the features
and the outcome. For each feature, we estimate its relevance to predicting the
outcome, and use these relevance estimates to rescale the feature matrix. This
rescaling procedure is equivalent to the feature selection methods, adaptive
lasso (\cite{zou2006adaptive}) and nonnegative garotte
(\cite{breiman1995better}). In the adaptive lasso, the adaptive scaling
factors are usually obtained from the ordinary least squares estimate. By
contrast, we inferred these scaling factors from an expert text corpus. This
technique leverages auxiliary data and can thus be applied when the original
training data are not sufficient to obtain reliable least squares estimates.
Other approaches to feature selection for predicting disease onset include
augmenting expert-derived risk factors with data-driven variables
(\cite{Sun2012}) and performing regression in multiple dimensions
simultaneously (\cite{wang2014clinical}).

\section{Data \& Features}
\vspace{-1mm}

\subsection{Data}
\vspace{-1mm}

We used the Taiwan National Health Insurance research database, longitudinal
cohort $3^{rd}$ release (\cite{hsiao2007using}), which contains billing
records for one million patients from 1996 to 2012. Because the dataset used a
Taiwan-specific diagnosis code (A-code) before 2000 instead of a more standard
coding system, we only utilized data after 2000.

We will detail the outcomes studied in the experiments section. For each
outcome, we used five years of data (2002-2007) to extract features and the
next five years (2007-2012) to define the presence of the outcome. We used
five-year periods based on the duration of available data and because
cardiovascular outcomes within five years are predicted by some risk
calculators (\cite{lloyd2010cardiovascular}).

\vspace{-1mm}
\subsection{Feature Transformation}
\vspace{-1mm}

For our purposes, the raw billing data was a list of tuples $(date,billing\
code)$ for each patient. Because we extracted features from five years of
data, each patient may have had multiple claims for any given billing code.
Let the \emph{count} of claims for patient $i$, billing code $j$ be $z_{ij}$.
The method of representing these counts ($z_{ij}$) in the feature matrix
($x_{ij}$) significantly impacts the prediction performance. We tried a number
of transformations and found that using the log function performed the best:
$x_{ij} = 1+log(z_{ij})\ if\ z_{ij}>0$, and $0$ otherwise. We normalized the
log-transformed value for each feature to $[0,1]$.

\vspace{-1mm}
\subsection{Features and Hierarchies}
\vspace{-1mm}

\begin{table}
\centering
\caption{Examples of Billing Codes.}


\label{w2v_feature_hierarchy}
\centering
\resizebox{\textwidth}{!}{
\begin{tabular}{@{} *{5}{l} @{}}
\hline
                                                & Description of           & Example & Example                                     & \# of  \\
Category                                        & hierarchy level          & code    & description                                 & codes  \\ \hline
\multirow{4}{25mm}{\textbf{Diagnoses (ICD-9)}}  & Groups of 3-digit ICD-9  & 390-459 & Diseases of the circulatory system          & 160    \\
                                                & 3-digit ICD-9            & 410     & Acute myocardial infarction                 & 1,018  \\
                                                & 4-digit ICD-9            & 410.0   & (as above) of anterolateral wall            & 6,442  \\
                                                & 5-digit ICD-9            & 410.01  & (as above) initial episode of care          & 10,093 \\ \hline
\multirow{4}{25mm}{\textbf{Procedures (ICD-9)}} & Groups of 2-digit ICD-9  & 35-39   & Operations on the cardiovascular system     & 16     \\
                                                & 2-digit ICD-9            & 35      & Operations on valves and septa of heart     & 100    \\
                                                & 3-digit ICD-9            & 35.2    & Replacement of heart valve                  & 890    \\
                                                & 4-digit ICD-9            & 35.21   & Replacement of aortic valve w/ tissue graft & 3,661  \\ \hline
\multirow{5}{25mm}{\textbf{Medications (ATC)}}  & Anatomical main group    & C       & Cardiovascular system                       & 14     \\
                                                & Therapeutic subgroup     & C03     & Diuretics                                   & 85     \\
                                                & Pharmacological subgroup & C03D    & Potassium-sparing agents                    & 212    \\
                                                & Chemical subgroup        & C03DA   & Aldosterone antagonists                     & 540    \\
                                                & Chemical substance       & C03DA01 & Spironolactone                              & 1,646  \\ \hline

\end{tabular}}
\end{table}

Our features consisted of age, and billing codes that were either diagnoses
and procedures (International Classification of Diseases $9^{th}$ Revision,
ICD-9), or medications (mapped to Anatomical Therapeutic Chemical, ATC). ICD-9
is the most widely used system of coding diagnosis and procedures in billing
databases (\cite{Cheng2011}), and ATC codes are used by the World Health
Organization to classify medications. Both coding systems are arranged as a
forest of trees, where nodes closer to the leaves represent more specific
concepts. Examples of feature descriptions and the potential number of
features are listed in Table~\ref{w2v_feature_hierarchy}. We used these
descriptions to estimate relatedness between each feature and the outcome.

\cite{Singh2014} showed that features that exploit the ICD-9 hierarchy improve
predictive performance. One of their methods was called ``propagated binary'':
all nodes are binary, and a node is 1 if that node or any of its descendants
are 1. In our modified ``propagated counts'' approach, the count of billing
codes at each node was the sum of the counts from itself and all of its
descendants. The log transformation was applied after the sum.

We removed all features that were present in less than three patients in each
training set, leaving approximately $10,000$ features for each outcome. The
first feature was patient's age in 2007, and the other features were related
to the counts of billing codes in the five years before 2007 and corresponded
to diagnoses, procedures, or medications.

\vspace{-1mm}
\section{Methods}
\vspace{-1mm}

\subsection{Computing Estimated Feature-Relevance}
\vspace{-1mm}

Figure~\ref{w2v_overview} summarizes our approach. We start with a word2vec
model trained on a corpus of medical articles, and use this model to compute
the relevance of each feature to the outcome. We then use this relevance to
rescale each feature such that the machine learning algorithm places more
weight on features that are expected to be more relevant.

\vspace{-1mm}
\subsubsection{Word2vec Model}
\vspace{-1mm}

\begin{figure}[!tb]
\centering
\includegraphics[width=4.5in]{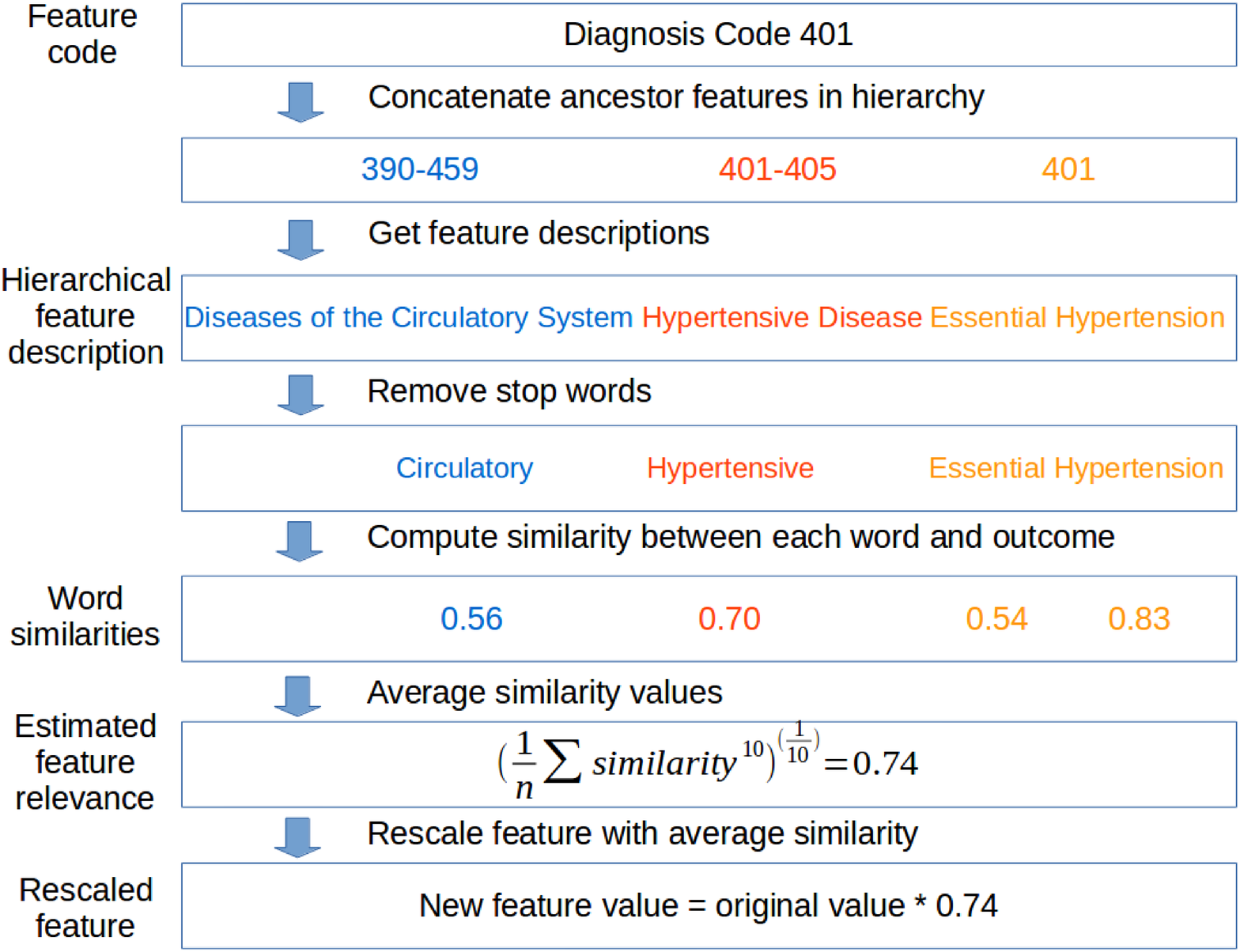}
\vspace{-5mm}

\caption{Example of computing and using the estimated relevance of each
feature for predicting diabetes.}

\label{w2v_feature_relevance}
\end{figure}

Word2vec uses a corpus of text to produce a low dimensional dense vector
representation $\in\Re^{d}$ for each word (\cite{mikolov2013efficient}). Words
used in similar contexts are represented by similar vectors, and these vectors
could be used to quantify similarity. We used models pre-trained on
$22,723,471$ PubMed article titles and abstracts and $672,589$ PubMed Central
Open Access full text articles (\cite{pyysalo2013distributional}). These
models used a skip-gram model with $d=200$ dimensional vectors and a window
size of 5. We used the learned word2vec vectors to estimate the relevance of
each features to predicting the outcome. The following sections detail this
process, and an example of this procedure applied to a single feature can be
found in Figure~\ref{w2v_feature_relevance}.

\vspace{-1mm}
\subsubsection{Construction of Hierarchical Feature Descriptions}
\vspace{-1mm}

Some feature descriptions are not informative without knowledge of its
ancestors, \emph{e.g.}, ICD-9 code 014.8 has the description ``others.'' To
ensure that each feature was assigned an informative description, we
concatenated each feature description with the descriptions of its ancestors
in the hierarchy. For example, diagnosis code 401 had as its ancestor the node
representing the group of codes 401-405, which in turn had as its ancestor
390-459. We defined the hierarchical feature description for code 401 as the
concatenated description for these three nodes
(Figure~\ref{w2v_feature_relevance}).

\vspace{-1mm}
\subsubsection{Computing Similarity for Each Word in Feature Description}
\vspace{-1mm}


We first removed extraneous or ``stop words'' based on a standard English list
(\cite{porter1980algorithm}), augmented with the words ``system,''
``disease,'' ``disorder,'' and ``condition.'' These augmented words were
chosen based on frequently occurring words in our feature descriptions. For
example, the original hierarchical description for code 401 (``Diseases of the
Circulatory System Hypertensive Disease Essential Hypertension'') was filtered
such that the remaining words were ``circulatory,'' ``hypertensive,''
``essential,'' and ``hypertension.'' We next applied the Porter2 stemmer
(\cite{porter1980algorithm}) to match each word in the descriptions to the
closest word in the word2vec model despite different word endings such as
``es'' and ``ed.''

Next, we computed the similarity between each word in the filtered
hierarchical feature description and the outcome description. For simplicity,
each outcome was summarized as a single keyword, e.g., ``diabetes''. In the
example for code 401, we computed the similarity between ``circulatory'' and
``diabetes'' by first extracting the two vector representations and taking the
cosine similarity. This similarity lies in the range $[-1,1]$. To obtain non-
negative scaling factors for our next step, we rescaled the similarity to
$[0,1]$. Finally we repeated this procedure for each of the remaining words.
In the example for code 401, we obtained four values: 0.56, 0.70, 0.54, and
0.83 (Figure~\ref{w2v_feature_relevance}).

\vspace{-1mm}
\subsubsection{Averaging Word Similarities}
\vspace{-1mm}

In each hierarchical feature description, the presence of a single word may be
sufficient to indicate the feature's relevance. For example, a feature that
contains the word ``hypertension'' in its description is likely to be relevant
for predicting diabetes. Mathematically, this would be equivalent to taking
the maximum of the values in the similarity vector. However, the maximum
function prevents relative ranking between features that contain the same
word. For example, ``screening for hypertension'' and the actual diagnosis
code for hypertension will receive the same relevance estimate although the
former description does not actually indicate the presence of hypertension.
Moreover, the maximum function is sensitive to outliers, and thus features
may be assigned erroneously high relevances.

By contrast, the arithmetic mean of the similarity vector reduces the effect
of any words that truly indicate high relevance. Thus, we selected the power
mean, which is biased towards the maximum, but takes all the words into
account. The power mean is defined as $(\frac{1}{m}\sum_i s_i^p)^\frac{1}{p}$,
where $s_i$ is the $i^{th}$ similarity value out of $m$ filtered words in the
hierarchical feature description. $p$ is a tunable exponent that can
interpolate the function between the maximum ($p=+\infty$) and the arithmetic
mean ($p=1$).

\begin{figure}[!tb]
\centering
\includegraphics[width=3.5in]{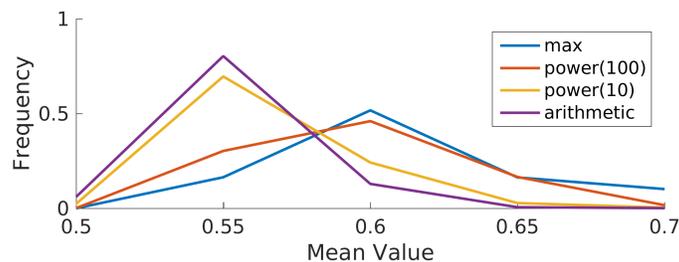}
\vspace{-5mm}

\caption{Distributions of feature relevances computed using various means:
arithmetic mean, power means with exponents 10 and 100, and maximum values.}

\label{w2v_means}
\end{figure}

We chose the exponent $p$ by plotting the distributions of feature relevances
when computed using various $p$ (Figure~\ref{w2v_means}). The $p=100$ curve
tracks the maximum closely, and may be overly sensitive to the maximum value
in each similarity vector, whereas the $p=10$ curve ``pushes'' an additional
10\% of the features towards higher relevances relative to the arithmetic
mean. Because we expected relatively few features to be relevant to the
outcome, we selected $p=10$. For example, the power mean of the word
similarities for code 401 was 0.74 (Figure~\ref{w2v_feature_relevance}).

\vspace{-1mm}
\subsubsection{Rescaling Features based on Similarity}
\vspace{-1mm}

Finally, we multiplied each feature value by the feature's estimated
relevance. This changed the numerical range of the feature from $[0,1]$ to
$[0,relevance]$. This is equivalent to using the adaptive lasso
(\cite{zou2006adaptive}) with an adaptive weight of $1/relevance$.

\vspace{-1mm}
\section{Experiment Setup}
\vspace{-1mm}

\subsection{Outcomes}
\vspace{-1mm}

We applied our method to predicting the onset of various cardiovascular
diseases. We focused on five adverse outcomes: cerebrovascular accident (CVA,
stroke), congestive heart failure (CHF), acute myocardial infarction (AMI,
heart attack), diabetes mellitus (DM, the more common form of diabetes),
hypercholesterolemia (HCh, high blood cholesterol).

For each outcome, we used five years of data before 2007 to predict the
presence of the outcome in the next five years. Although billing codes can be
imprecise in signaling the presence of a given disease, we reasoned that
repeats of the same billing code should increase the reliability of the label.
Thus based on advise from our clinical collaborators, we defined each outcome
as three occurrences of the outcome's ICD-9 code or the code for the
medication used to treat the disease. To help ensure that we were predicting
the onset of each outcome, we excluded patients that had at least one
occurrence of the respective billing codes in the first five years.





We built separate models for two age groups: (1) ages 20 to 39, and (2) ages
40 to 59 (Table~\ref{w2v_populations}) and for males and females. We separated
the age groups because the incidence of the outcomes increased with age. The
age 40 was chosen to split our age groups because 40 is one of the age cutoffs
above which experts recommend screening for diabetes
(\cite{siu2015screening}). Patients below 20 and above age 60 were excluded
because of very low or high rates of these outcomes, respectively. In
addition, we built separate models for males and females because many
medications and diagnoses are strongly correlated with gender, potentially
confounding the interpretation of our models. In summary, we built models for
five diseases, two genders, and two age groups for a total of 20 prediction
tasks.

\begin{table}
\centering

\caption{Prediction tasks and number of patients. CVA= cerebrovascular
accident (stroke); CHF= congestive heart failure; AMI= acute myocardial
infarction; DM= diabetes mellitus; HCh= hypercholesterolemia.}

\label{w2v_populations}
\centering
\vspace{-3mm}
\begin{tabular}{@{} {l} *{4}{c} @{}}
\hline
Outcome & \multicolumn{2}{c}{Age group 1: $20-39$} & \multicolumn{2}{c}{Age group 2: $40-59$} \\
-gender    & N      & n (\%)        & N      & n (\%)         \\\hline
CVA-F     & 171836 &  100  (0.1\%) & 141222 &  1045  (0.7\%) \\
CVA-M     & 156649 &  224  (0.1\%) & 138210 &  2055  (1.5\%) \\\hline
CHF-F     & 171749 &  137  (0.1\%) & 141366 &   838  (0.6\%) \\
CHF-M     & 156704 &  217  (0.1\%) & 139276 &  1056  (0.8\%) \\\hline
AMI-F     & 170247 &  552  (0.3\%) & 131780 &  4507  (3.4\%) \\
AMI-M     & 154921 & 1109  (0.7\%) & 129177 &  6088  (4.7\%) \\\hline
DM-F      & 168194 & 1609  (1.0\%) & 128715 &  6879  (5.3\%) \\
DM-M      & 153834 & 2273  (1.5\%) & 125126 &  8609  (6.9\%) \\\hline
HCh-F     & 166938 & 2386  (1.4\%) & 117531 & 13203 (11.2\%) \\
HCh-M     & 149061 & 5008  (3.4\%) & 113839 & 13874 (12.2\%) \\\hline

\end{tabular}
\end{table}

\subsection{Experimental Protocol}
\vspace{-1mm}

For each prediction task, we split the population of patients into training
and test sets in a 2:1 ratio, stratified by the outcome. We learned the
weights for a L1-regularized logistic regression model on the training set
using liblinear (\cite{Fan2008}). The cost parameter was optimized by two
repeats of five-fold cross validation. Because only $0.1\%$ to $12\%$ of the
population experienced the outcome, we set the asymmetric cost parameter to
the class imbalance ratio. We repeated the training/test split 10 times, and
report the area under receiver operating characteristic curve (AUC)
(\cite{Hanley1982}) averaged over the 10 splits.

To assess the statistical significance of differences in the AUC for each
task, we used the sign test, which uses the binomial distribution to quantify
the probability that at least $n$ values in $m$ matched pairs are greater
(\cite{rosner2015fundamentals}). This requires no assumptions about the
distribution of the AUC over test sets. When two methods are compared,
$p<0.05$ is obtained when one method is better in at least 9 out of the 10
test sets. In our experiments, we compare the performance of models that use
our rescaling procedure with ones that do not (the ``standard'' approach).

\vspace{-1mm}
\section{Experiments and Results}
\vspace{-1mm}

\subsection{Ranking of Features}
\vspace{-1mm}

We first verified that the word2vec similarity and our power law averaging
provided a sensible relative ranking of features with respect to each outcome.
For the AMI (heart attack) outcome for example, highly ranked features
descriptions were related to the heart, with a similarity ranging from $0.65$
to $0.85$. The least similar feature descriptions, such as skin disinfectants
and cancer drugs, had similarities close to $0.5$.

\subsection{Full Dataset}
\vspace{-1mm}

Because there was a wide range in number of outcomes in our tasks, we reasoned
that evaluating our method on the full dataset would provide information about
the regime of data size for which our method might be useful. Thus, we first
trained models for each of the 20 tasks and assessed their performance. Our
proposed method meaningfully outperformed the standard approach in one task
(AUC of $0.718$ compared to $0.681$ for CVA in age group 1 of females). We had
the fewest number of positive examples in the training data (67) for this
task. Any statistically significant differences in the remaining tasks were
small, ranging from $-0.009$ to $0.009$. This indicated that our method had
minimal effect on performance (either positive or negative) when data were
plentiful. In 13 out of 16 tasks where the number of selected features were
statistically different, our method selected fewer features. On average, our
method selected 83 features, compared to 117 for the standard approach.

Because the most significant difference and several non-statistically
significant differences occurred in predictive tasks with fewer than 100
outcomes in the training set, we hypothesized that our method would be most
useful in situations with few positive examples.

\begin{figure}[!tb]
\centering
\includegraphics[width=5.5in]{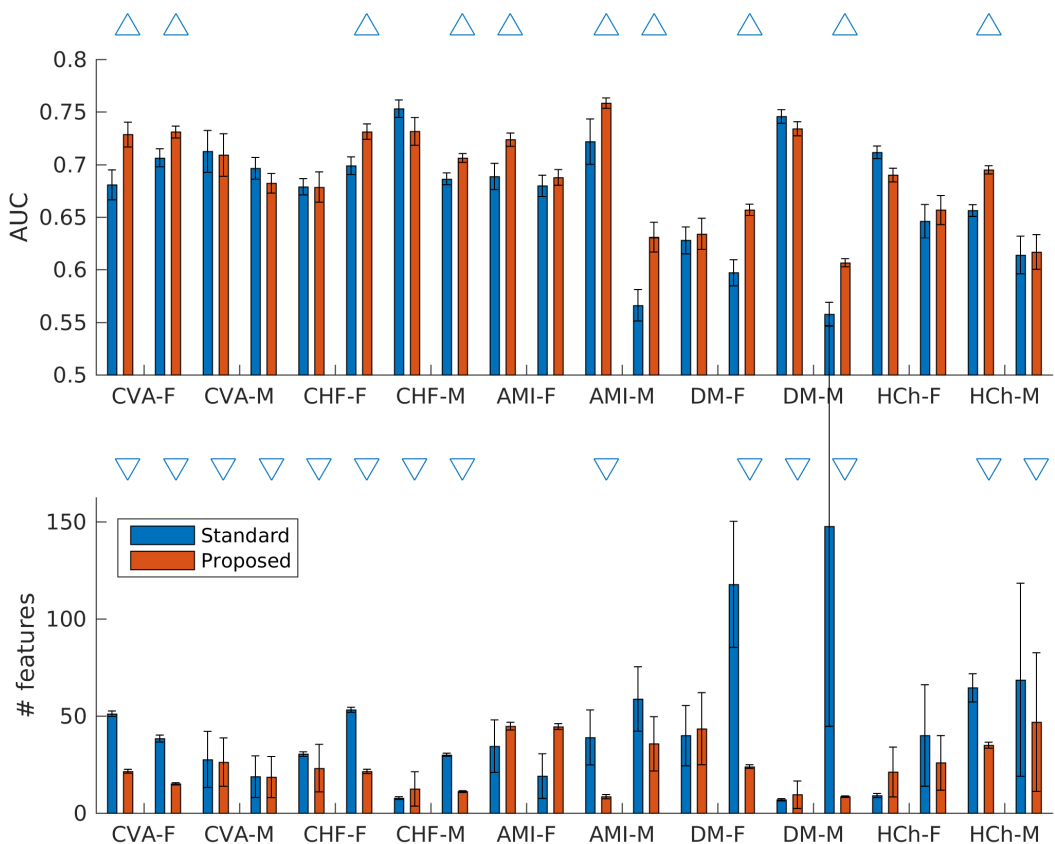}
\vspace{-5mm}

\caption{AUC and number of selected features using 50 positive examples using
the standard approach compared with our proposed rescaling approach. Outcome
and gender are labeled on the x-axis, each spanning two predictive tasks: age
groups 1 and 2. Triangles indicates a significantly higher (pointed up) or
lower (pointed down) AUC or number of selected features for our proposed
method. Error bars indicate standard error.}

\label{w2v_results_ds}
\end{figure}

\vspace{-1mm}
\subsection{Downsampled Dataset with 50 Positive Examples}
\vspace{-1mm}

To test our hypothesis, we used the same splits as in the previous section,
and downsampled each training split such that only 50 positive examples and a
proportionate number of negative example remained. We assessed the effect of
training on this smaller training set, and computed the AUC on the same
(larger) test sets.

Our proposed method yielded improvements relative to the standard approach in
10 out of 20 tasks, with differences ranging from $0.020$ to $0.065$
(Figure~\ref{w2v_results_ds}). The remaining 10 tasks did not have
statistically significant differences. Among the tasks with significant
differences, our method had an average AUC of $0.697$, compared to $0.656$ for
the standard approach. In addition, where the number of selected features were
statistically different, our method selected 20 features, compared to 50 for
the standard approach.

\vspace{-1mm}
\subsection{Downsampled Dataset with 25 Positive Examples}
\vspace{-1mm}

After we further downsampled the training data to have 25 positive examples,
similar observation were made: our method yield improvements in 6 tasks, with
differences ranging from $0.022$ to $0.086$. The remaining 14 tasks did not
have statistically significant differences. Among the tasks with significant
differences, our method had an average AUC of $0.658$ compared to $0.606$ for
the standard approach. Similarly, where the number of selected features were
statistically different, our method selected 52 features, compared to 80 for
the standard approach.

\begin{figure}[!tb]
\centering
\includegraphics[width=2.5in]{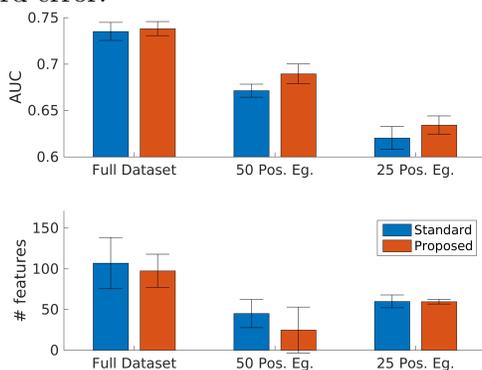}
\vspace{-5mm}

\caption{AUC and number of selected features in the full and downsampled
datasets averaged across all prediction tasks.}

\label{w2v_results_summary}
\end{figure}

A summary of results from all three experiments are in
Figure~\ref{w2v_results_summary}, showing the overall trend of improved
performance and fewer selected features when comparing our method to the
standard approach.

\vspace{-1mm}
\section{Discussion}
\vspace{-1mm}

We demonstrated a method to leverage auxiliary, publicly available free text
data to improve prediction of adverse patient outcomes using a separate,
structured dataset. Our method improved prediction performance, particularly
in cases of small data. Furthermore, our method consistently selects fewer
features (20 vs 50) from an original feature dimensionality of $10,000$. This
allows domain experts to more easily interpret the model, which is a key
requirement for medical applications to see real world use. To our knowledge
this is the first work that transfers knowledge in this manner.

\begin{figure}[!tb]
\centering
\includegraphics[width=4.5in]{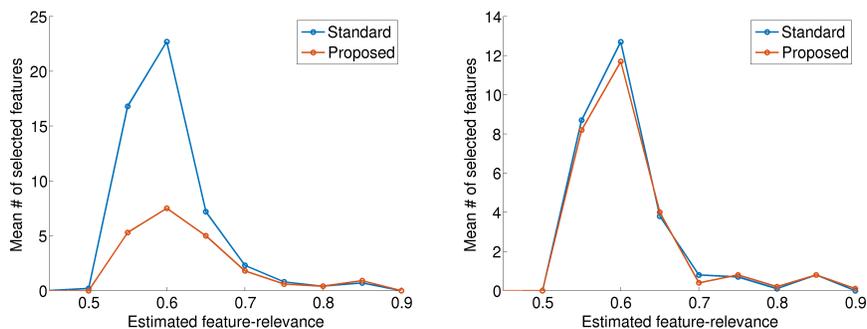}
\vspace{-5mm}

\caption{Number of selected features by estimated feature-relevance for
predicting stroke in patients age 20-39 in females (left) and males (right).}

\label{w2v_stroke_selected_relevance}
\end{figure}


To better understand the reason for our improved performance and number of
selected features, we plotted the number of selected features against the
estimated feature-relevance. The plots for predicting stroke in patients aged
20-39 are shown in Figure~\ref{w2v_stroke_selected_relevance}: females on the
left and males on the right. In females, where the method improved
performance, there is a marked decreased in the number of selected features
with estimated relevance below 0.7, but little change among features with
higher estimated relevances. In other words, our approach preferentially
removed features with low estimated relevance. Removal of these ``noise''
features improved performance.

Diagnosis code 373.11 (Hordeolum externum) is an example of a feature that was
selected by the original approach but not our method. This condition, also
called a sty, is a small bump on the surface of the eyelid caused by a clogged
oil gland. Our approach assigns this feature a relevance of 0.57, which
decreased its probability of being selected.

In males however, where neither the performance nor the number of selected
features were meaningfully different, there was little difference in the
estimated relevance of selected features. These plots were representative of
other tasks where the performance was statistically different or unchanged.

An advantage of our method is its small number of parameters; the power mean
exponent is the only tunable parameter, and can be selected by cross
validation. Because our method is not specific to medicine, it can also be
applied to other domains where feature descriptions are available (as opposed
to numerically labeled features). In the absence of a corresponding domain
specific text corpus, a publicly available general corpus such as Wikipedia
may suffice to estimate the relevance of each feature to the outcome.

\vspace{-1mm}
\section{Limitations \& Future Work}
\vspace{-1mm}

Our work has several limitations. First, the data are comprised of billing
codes, which may be unreliable for the purposes of defining the presence or
absence of a disease. Unfortunately, measurements such as blood sugar or A1C
for diabetes were not available in our dataset.

Our method of estimating relevance can also be improved by training a separate
model to predict the relevances using inter-feature correlations in the data.
This may handle issues such as differences in patient population (e.g., age
group and gender), and syntax such as negation in the feature descriptions.

\vspace{-1mm}
\bibliography{transfer_w2v}

\end{document}